%% file: arxiv.tex
\documentclass{article} 
\usepackage[T1]{fontenc}
\usepackage{iclr2027_conference,times}
\input{math_commands.tex}

\usepackage{xcolor}
\usepackage{colortbl}
\definecolor{navyblue}{RGB}{11,26,58}
\definecolor{nvgreen}{RGB}{136,204,2}

\definecolor{resultfirst}{RGB}{174,214,192}
\definecolor{resultsecond}{RGB}{208,232,220}
\definecolor{resultthird}{RGB}{234,244,238}
\newcommand{\rankfirst}[1]{\cellcolor{resultfirst}#1}
\newcommand{\ranksecond}[1]{\cellcolor{resultsecond}#1}
\newcommand{\rankthird}[1]{\cellcolor{resultthird}#1}
\newcommand{\resultranklegend}{\par\vspace{3pt}{\scriptsize
  \setlength{\fboxsep}{1pt}
  \colorbox{resultfirst}{Best}\quad\colorbox{resultsecond}{Second best}\quad
  \colorbox{resultthird}{Third best}\quad By metric; ties share a shade.\par}}
\newcommand{\textranklegend}{\par\vspace{3pt}{\scriptsize
  \textbf{Best}\quad\underline{Second best}
  \quad By metric; ties share formatting.\par}}
\usepackage{url}
\usepackage{amsmath,amssymb,amsthm}
\usepackage{booktabs}
\usepackage{multirow}
\usepackage{graphicx}
\usepackage{adjustbox}  
\usepackage{seqsplit}
\usepackage{float}
\usepackage{flafter} 
\usepackage[section,below]{placeins} 
\usepackage{wrapfig}
\usepackage{enumitem}
\usepackage{algorithm}
\usepackage{algpseudocode}

\usepackage{titletoc}

\usepackage{xspace}
\usepackage[colorlinks=true,
            linkcolor=nvgreen,
            citecolor=nvgreen,
            urlcolor=nvgreen]{hyperref}
\newcommand{\ourmethod}{\textsc{CtrlWAM}\xspace}

\newcommand{\act}{\mathbf{a}}   
\newcommand{\vid}{\mathbf{v}}   
\newcommand{\x}{\mathbf{x}}     
\newcommand{\render}{\mathbf{r}}

\title{\raggedright\textcolor{nvgreen}{Ctrl}\textcolor{nvgreen}{WAM}: Controllable World Action Models with Aligned Intent and Foresight}
\newcommand{\website}{\url{https://ctrl-wam.github.io/}}

\author{\textbf{Chensheng Peng}$^{1,2*}$ \qquad \textbf{Wenhao Ding}$^{1}$ \qquad
\textbf{Ran Tian}$^{1}$ \qquad \textbf{Zewei Zhou}$^{1,3*}$ \\
\textbf{Jef Packer}$^{1}$ \qquad \textbf{Maximilian Igl}$^{1}$ \qquad
\textbf{Peter Karkus}$^{1}$ \qquad \textbf{Yan Wang}$^{1}$ \\
\textbf{Masayoshi Tomizuka}$^{1}$ \qquad \textbf{Boris Ivanovic}$^{1}$ \qquad
\textbf{Marco Pavone}$^{1}$ \qquad \textbf{Yuxiao Chen}$^{1}$ \\[0.5em]
$^{1}$NVIDIA \qquad
$^{2}$UC Berkeley \qquad
$^{3}$UCLA \\[0.18cm]
}

\iclrfinalcopy 
\begin{document}

\maketitle

\begin{figure}[H]  
\centering
\vspace{-1cm}
\includegraphics[width=\linewidth]{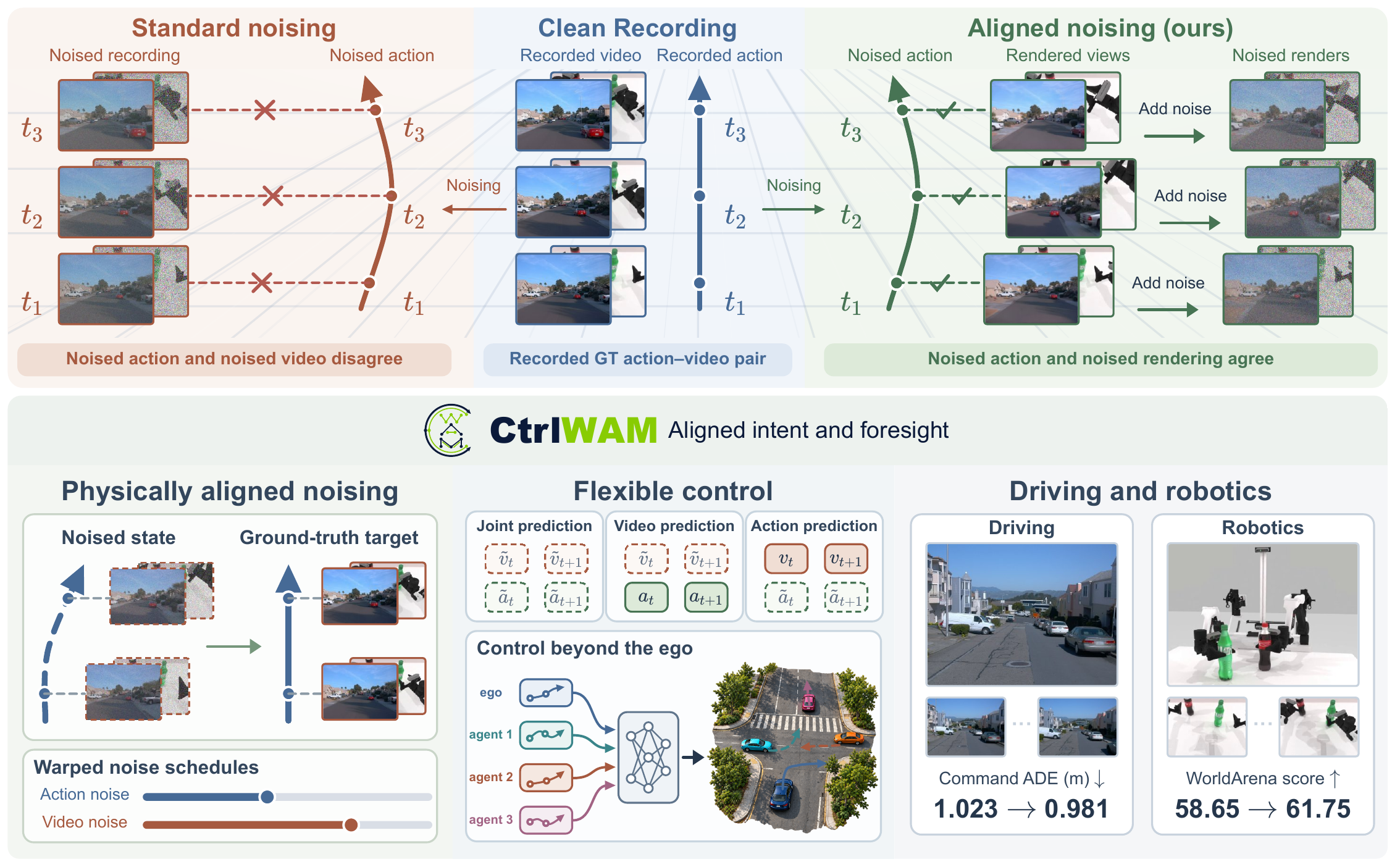}
\vspace{-0.8cm}
\caption{ Overview of \ourmethod. Standard noising pairs perturbed actions with video tied to the recorded future. Aligned noising instead renders the consequences of the perturbed trajectories before adding visual noise, while retaining the recorded action–video pair as the training target. Combined with warped video–action noise schedules, the model supports joint prediction, action-conditioned video generation, video-conditioned action prediction, and control of surrounding agents, with applications to driving and robotic manipulation.}
\label{fig:teaser}
\end{figure}

\renewcommand*{\thefootnote}{\fnsymbol{footnote}}
\footnotetext{
    \raggedright
    $^{*}$Work done during internship at NVIDIA}

\begin{abstract}
World action models (WAMs) jointly predict actions (intent) and visual future (foresight). Standard training adds noise to recorded actions and video simultaneously, but such training paradigms introduce a mismatch: perturbed actions imply counterfactual future visual, while the noised video remains tied to the GT recording. In low-noise regime, the scene geometry and even the dynamic behavior remain clearly visible from the noisy future frames despite the added noise.
We present \ourmethod, which executes perturbed actions in a simulator and pairs them with their noised visual consequences for joint WAM learning.
To accommodate the different denoising requirements of video and actions, we introduce warped video--action noise schedules that aim to keep visual layout responsive as action predictions evolve. 
We further extend the action interface from ego-only control to a variable number of agent streams, allowing a unified model to represent predicted or commanded futures for multiple agents. 
Driving experiments show more accurate action forecasts, closer agreement between generated video and actions, and better following of supplied commands; robotics experiments show stronger motion fidelity and controllability. Matched controls support the benefit of off-path renders for command following and manipulation fidelity. Together, these findings contribute to a more controllable world action model. Project page: \website
\end{abstract}

\section{Introduction}
\label{sec:introduction}
World action models (WAMs) jointly predict an embodied agent's future actions (\emph{intent}) and their visual consequences (\emph{foresight})~\citep{zhu2025uwm,ye2026world}.
The imagined video should therefore remain consistent with the consequences of predicted actions.
In unified models such as Cosmos~3~\citep{agarwal2026cosmos3}, which perform both joint prediction and forward dynamics (action-conditioned video generation), this intent--foresight alignment is essential in both modes.

\begin{wrapfigure}{r}{0.42\textwidth}
\centering
\includegraphics[width=\linewidth]{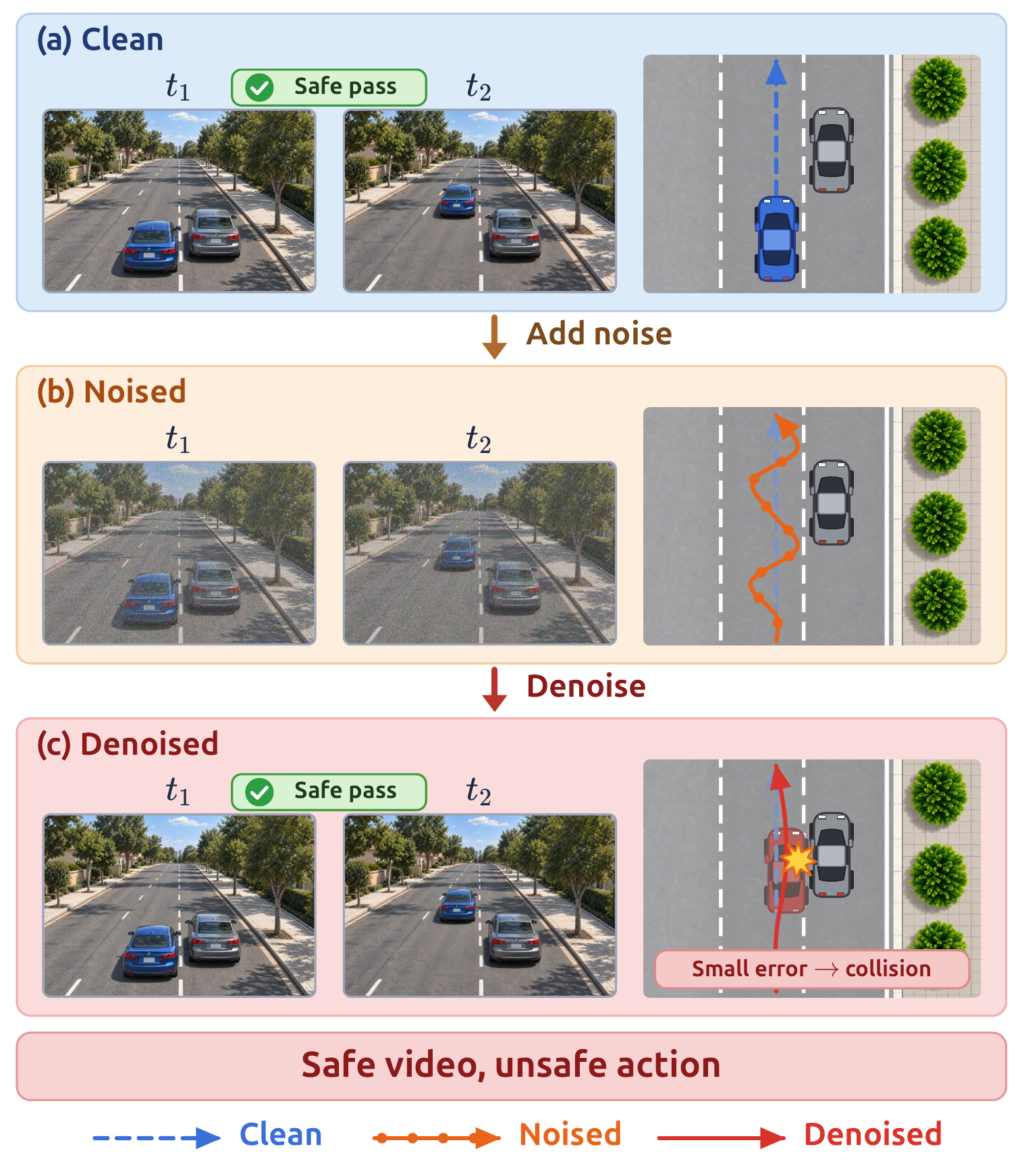}
\vspace{-20pt}
\caption{\textbf{Safe video, unsafe action.} This schematic illustrates intent–foresight misalignment: after noising and denoising, the video depicts a safe pass, while a small deviation in the predicted trajectory leads to a collision.}
\label{fig:moti}
\end{wrapfigure}

However, current training procedures do not explicitly preserve this alignment during denoising.
Methods such as DreamZero~\citep{ye2026world} independently corrupt the recorded actions and video with Gaussian noise~\citep{ho2020denoising}.
Perturbing the actions in this way does not change the motion depicted in the video: the corrupted video remains tied to the recorded future, even when the perturbed actions would produce a different trajectory.
The model is thus trained to recover the paired clean targets, but it receives no explicit supervision on the physical correspondence between its intermediate action and video states.

This gap matters because useful visual structure can emerge early in denoising.
DIDO~\citep{lyu2026dido} observes that static scene structure forms before gripper--object dynamics are fully resolved.
In its four-step video-model ablation, conditioning the action expert on first-step representations already achieves 97.7\% success on LIBERO, compared with 98.4\% after all four steps.
These findings suggest that intermediate visual predictions already provide substantial information for action generation, which motivates closer attention to their physical consistency with the evolving action estimates.
Without explicit alignment, an intermediate action estimate could veer toward an obstacle while the imagined video still depicts a safe pass (Figure~\ref{fig:moti}).

We introduce \ourmethod for physically aligned denoising in world action models.
Its central idea, \emph{aligned noising}, pairs each action perturbation with its visual consequence to construct physically corresponding training inputs (Figure~\ref{fig:teaser}).
Specifically, a simulator executes the perturbed actions and renders the resulting observations, and we use RoboTwin~\citep{chen2025robotwin} for manipulation and AlpaSim~\citep{alpasim_2025} for driving.
We then add noise to the rendered video and train the model to recover the recorded actions and video at the corresponding timestamps.
These paired inputs provide a physically grounded training signal for flow matching~\citep{lipman2023flow}.

Learning from these pairs also requires choosing how much visual evidence to expose at each stage of denoising.
Video and actions can be assigned separate noise levels~\citep{chen2024diffusionforcing} and can therefore follow different noise schedules~\citep{guo2026unified}.
Our preliminary inspection suggests that the video layout can stabilize and enter detail refinement while the action estimates are still changing (Figure~\ref{fig:denoising_misalignment}).
This observation motivates \emph{warped video--action noise schedules}.
By assigning video a higher noise level than actions, we aim to keep the video layout responsive to evolving action predictions, and then let it settle into detail refinement as the actions approach convergence.

In interactive scenes, the future also depends on the behavior of other agents. A driving scenario, for example, may leverage a neighboring vehicle's maneuver in addition to the history ego trajectory for the action prediction.
We therefore introduce \emph{structured multi-agent conditioning}, which extends the ego action stream to a variable number of additional agents.
Conditioning masks specify which futures are predicted and which are supplied as commands.
This interface supports both joint multi-agent prediction and action-conditioned generation, and it reduces exactly to the ego-only model when no other agents are present.

We evaluate \ourmethod on autonomous driving~\citep{alpasim_2025} and bimanual manipulation~\citep{chen2025robotwin}.
In driving, we assess action forecasting, video--action consistency, and adherence to supplied commands. In manipulation, we assess motion fidelity and video quality.

In summary, our main contributions are as follows:
\begin{itemize}[leftmargin=1.5em,itemsep=2pt]
\item \textbf{Physically aligned noising}, which pairs perturbed actions with their simulator-rendered visual consequences and trains the model to recover the recorded actions and video, providing physically corresponding inputs for video--action denoising.
\item \textbf{Warped video--action noise schedules}, which assign video a higher noise level than actions, aiming to keep the video layout adaptable while actions evolve and to shift toward detail refinement as they stabilize.
\item \textbf{Structured multi-agent conditioning}, which extends the ego action stream with streams for additional agents, so that each agent's future can be either predicted to improve the ego-prediction or supplied as a command to enhance the controllability.
\end{itemize}

\section{Related Work}
\label{sec:related}

\paragraph{Unified World Action Models}
Joint video--action models learn action prediction and visual dynamics within a shared generative model. The Unified World Models (UWM) framework couples video and action diffusion within one transformer \citep{zhu2025uwm}. The Unified Video Action (UVA) model learns a joint latent representation with separate video and action decoders \citep{li2025uva}. Both support policy learning and forward or inverse dynamics through flexible conditioning. VideoVLA and DreamZero adapt pretrained video generators for joint prediction of actions and visual futures in robotic control \citep{shen2025videovla,ye2026world,yuan2026fast}. 

\paragraph{Controllable World Models}
Controllable world models condition visual futures on commands, actions, or scene structure. In robotics, action-conditioned video prediction supports model-predictive planning \citep{finn2017foresight}, while Genie learns latent actions for interactive generation from unlabeled videos \citep{bruce2024genie}. GAIA-1 uses video, text, and action inputs to steer driving scenarios \citep{hu2023gaia}, while DriveDreamer learns structured traffic constraints for controllable generation \citep{wang2023drivedreamer}. Vista supports controls ranging from high-level commands and goal points to trajectories, steering angles, and speeds \citep{gao2024vista}. 


\section{\ourmethod}
\label{sec:method}

\subsection{Physically Aligned noising}
\label{sec:aligned-training}

A recorded action--video pair specifies how the two modalities agree at the clean endpoint. It does not specify how they should deviate together from that endpoint as noise is added.
For example, noising an action trajectory produces a different ego camera path, implying a different visual future, yet adding Gaussian noise to the recorded video does not reflect this change (Figure~\ref{fig:moti}).
Aligned noising instead constructs an action perturbation and its visual outcome as a paired training input (Figure~\ref{fig:teaser}).
This gives the model physically related evidence in both modalities as it learns to recover their shared recorded future.

For each recording, we sample $t\in[0,1)$, which sets the action noise level $\sigma_a=1-t$. We then construct the perturbed command $\act_t^{\mathrm{cmd}}=t\act^\star+(1-t)\eps_a$ with $\eps_a\sim\mathcal{N}(0,I)$.
A simulator executes this command and renders its visual consequences. We use RoboTwin~\citep{chen2025robotwin} for manipulation and AlpaSim~\citep{alpasim_2025} for driving.
We store the executed action trajectory $\act_t$ and the latents $\render$ of the rendered video as one training pair.
The recorded history provides the conditioning prefix, and the recorded future over the same timestamps as the rollout supplies the clean supervision targets.

The executed trajectory $\act_t$ serves as the perturbed action input to the denoiser. The video, in contrast, is corrupted with freshly sampled noise:
$\vid_{\sigma_v}=\sigma_v\eps_v+(1-\sigma_v)\render$ with $\eps_v\sim\mathcal{N}(0,I)$.
The video noise level $\sigma_v$ is assigned separately, as described in Section~\ref{sec:ias}.
The resulting input therefore pairs a perturbed action with a corrupted view of \emph{its} consequences, rather than with a noised view of the original, unperturbed future.
We refer to this pairing as \emph{physical correspondence}.

Keeping the recording as the target turns the simulated deviation into an input from which the model learns to recover a reference future. Using the perturbed rollout as the target would instead teach the model to reconstruct that alternative future, rather than the common future. 
Because the input no longer lies on the standard interpolation path, the denoising target accounts for the full difference between the actual training input and the recording. This difference includes both the change introduced by the simulated motion and the added video noise.
The model is thus supervised to correct both the behavioral and the visual discrepancy in each paired example.

\subsection{Warped video--action noise schedules}
\label{sec:ias}

\begin{figure}[!htbp]
\centering
\includegraphics[width=\linewidth]{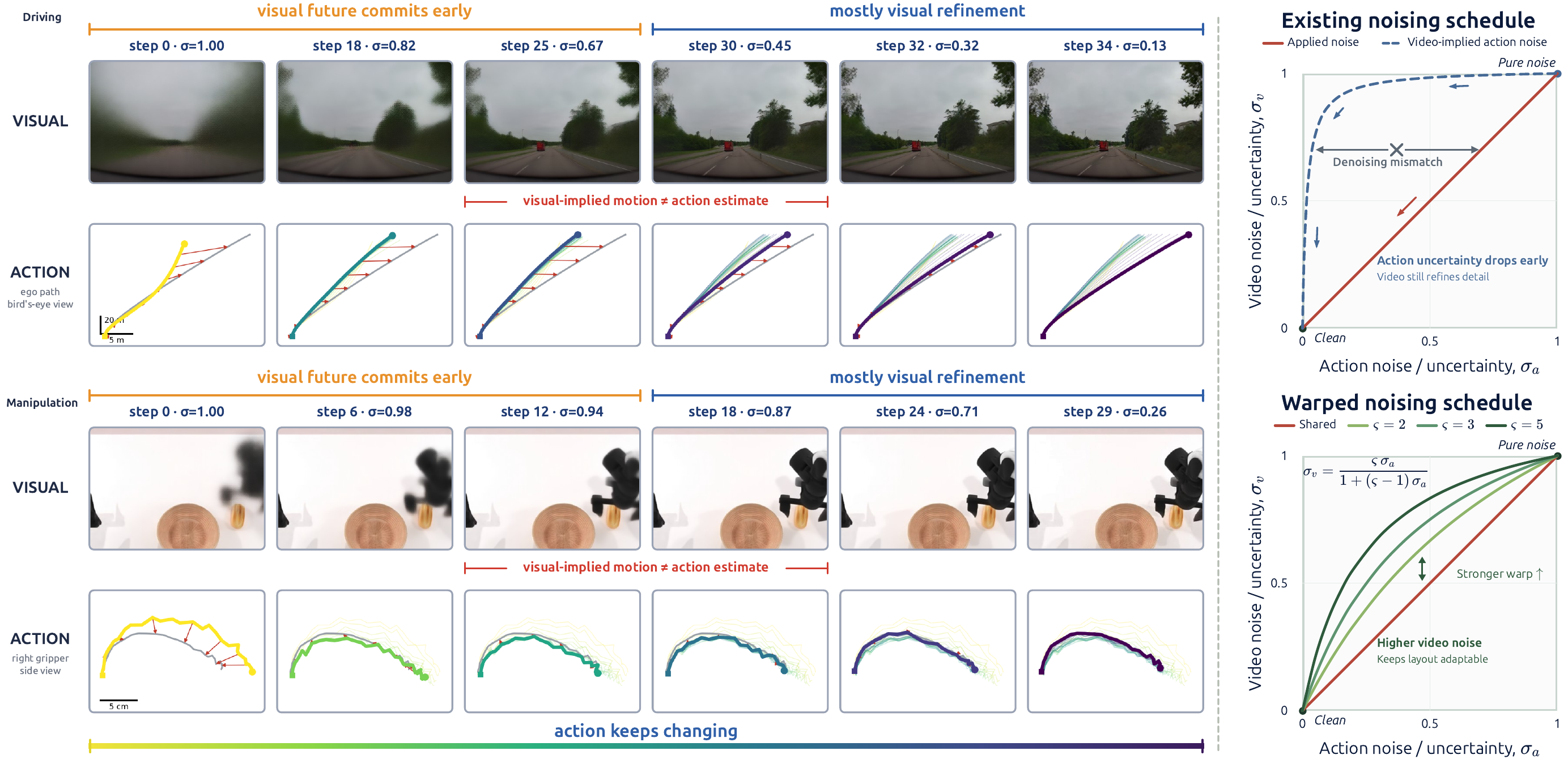}
\vspace{-0.5cm}
\caption{Visual and action predictions evolve at different rates during denoising. Left: driving and manipulation examples show that visual structure becomes established earlier, while the estimated ego and gripper trajectories continue to change. Columns indicate denoising steps and noise levels. Right: schematic schedules illustrate the mismatch under shared noising.  Our proposed warp instead assigns higher video noise at a given action noise level.
}
\label{fig:denoising_misalignment}
\end{figure}

During joint generation, the video can commit to a layout in early denoising steps while the action estimates are still changing.
Our preliminary inspection (Figure~\ref{fig:denoising_misalignment}) suggests that later video denoising steps then mainly refine appearance (left). It can be observed that the implied action from the video stabilizes quickly although the video is still on a high-level noise. Meanwhile, the explicitly predicted actions are still changing.
As a result, the motion implied by the early video may differ from that of the evolving action. A shared noise schedule therefore allows the video to commit to a trajectory before the action prediction settles, and there is a mismatch between the video-implied action and the predicted action.

Warped video--action noise schedules aim to pace this process. They keep the layout open to revision while the actions evolve, and then let it settle into detail refinement as the actions approach convergence.
We encourage this progression by assigning video a higher noise level than actions during training.
At a given action noise level, the model therefore sees a more strongly corrupted version of the corresponding render, which limits its reliance on an already clear visual layout.
Building on separate noise levels for different modalities~\citep{zhu2025uwm}, we adapt the timestep-shift formulation of \citet{esser2024sd3} and set
\begin{equation}
\sigma_v=\mathrm{warp}(\sigma_a;\varsigma)
=\frac{\varsigma\sigma_a}{1+(\varsigma-1)\sigma_a},
\label{eq:ias-warp}
\end{equation}
where $\varsigma$ controls the strength of the warp.

Setting $\varsigma=1$ recovers the shared schedule $\sigma_v=\sigma_a$.
For $\varsigma>1$, the video input contains more Gaussian noise and less rendered content at a fixed action noise level, and the two levels remain deterministically coupled.
Each predicted modality receives its own timestep embedding, and actions supplied as conditioning are kept fixed.

Along this training path, the video remains more corrupted while the action is uncertain, and both modalities approach their clean endpoints as the action noise level decreases.
The intended effect is to leave room for changes in scene layout before generation concentrates on fine detail.

\begin{figure}[!tbp]
\centering
\vspace{-0.5cm}
\includegraphics[width=\linewidth]{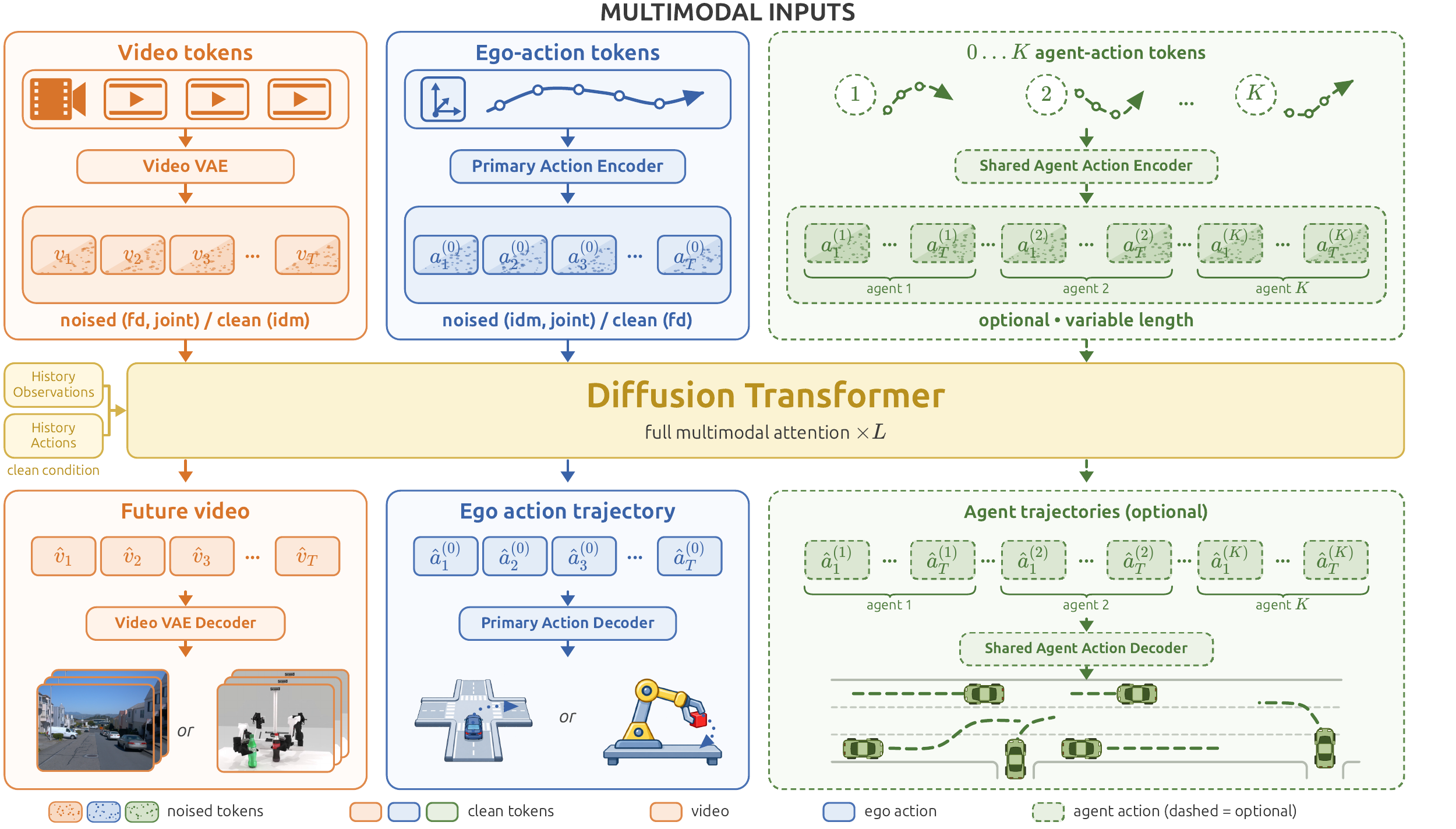}
\caption{The architecture of \ourmethod.  A shared diffusion transformer processes video, ego-action, and optional agent-action tokens, conditioned on observed history. Future tokens are denoised as prediction targets or supplied clean as conditions, enabling joint prediction, action-conditioned video generation, and video-conditioned action prediction. The additional agent streams support prediction and control of surrounding agents.}
\label{fig:model-architecture}
\end{figure}

\subsection{Structured multi-agent control}
\label{sec:multiagent-method}

Interactive driving would benefit from both predicting the behavior of surrounding agents and supporting explicit control over their actions.
Even when no commands are provided, the likely maneuvers of these agents shape the ego vehicle's decisions, so predicting them matters for modeling the ego trajectory.
The ego-only action interface leaves these futures implicit and provides no dedicated channel for commanding individual agents.
We therefore extend the shared transformer with additional agent streams that represent surrounding agents explicitly, supporting both joint behavior prediction and selective control.

The extended action input contains one ego block and $K$ agent blocks, each with history and future tokens (Figure~\ref{fig:model-architecture}).
When no future actions are supplied, the model jointly predicts the ego trajectory, the trajectories of the other agents, and the video.
The predicted agent futures serve as estimates of each agent's intent, so that ego-action prediction can account for anticipated motions and interactions.
When commands are available, the corresponding agent tokens are supplied clean, and the model predicts the video tokens.
A conditioning mask implements these roles. Noised future tokens are denoised as prediction targets, whereas clean future tokens act as commands.
In both prediction and controlled generation, the surrounding agents thus remain explicitly represented in the model.



\section{Experiments}
\label{sec:experiments}
\label{sec:results}


\subsection{Datasets and Evaluation Metrics}
\label{sec:data-evaluation}

\textbf{Driving datasets.}
We use recorded sequences from NVIDIA's PhysicalAI Autonomous Vehicles dataset \citep{nvidia2025physicalaiav} and 3DGS scene reconstructions from NuRec \citep{nvidia2025physicalaiavnurec} to generate perturbed renderings. Each recorded clip is around 20 seconds. We slice each clip into multiple short simulation windows over approximately 8,900 reconstructed scenes, yielding 33,988 usable clip windows after kinematic, encoding, and timestamp-pairing checks. The training split provides approximately 270,000 aligned items with up to eight perturbation levels per window. Appendix~\ref{app:data} provides the data inventory details.

\textbf{Driving evaluation.}
The held-out evaluation set contains 1,156 clips and 1,434 event-centered segments. We train a separate inverse dynamics model (IDM) to estimate trajectories from videos, implying the intent embedded in the video. In \emph{Joint mode}, the model predicts future video and actions together. In \emph{forward dynamics}, future actions are supplied as commands and the model predicts only video. The main benchmark uses recorded future actions as commands; the counterfactual tests replace them with modified trajectories. 

\textbf{Driving metrics.}
Using the standard ADE/FDE definitions ~\citep{gupta2018socialgan}, in Joint mode, \emph{Self ADE} compares the IDM read-out of the generated video with the model's own predicted trajectory.  It measures video--action consistency. \emph{Rollout ADE} compares the directly generated action trajectory with the recording and measures action forecasting accuracy without an IDM read-out. In forward dynamics, we report \emph{Cmd ADE} by comparing the IDM read-out from the generated video with the trajectory specified by the supplied action command. 
We also report future-frame peak signal-to-noise ratio (PSNR) \citep{hore2010psnr} for visual reconstruction quality.

\textbf{Robotics datasets.}
We use RoboTwin 2.0 \citep{chen2025robotwin} for bimanual manipulation. We curated 59,696 simulator-derived samples consisting of the perturbed action-rendering pairs. Generated videos are evaluated on 1,000 held-out episodes using WorldArena \citep{worldarena2025}.

\textbf{Robotics metrics.}
Our WorldArena Track-1 evaluation reports 15 metrics grouped into six dimensions: visual quality, motion quality, content consistency, physics adherence, 3D accuracy, and controllability \citep{worldarena2025}. EWMScore is $100$ times the mean of the 15 individual metrics. 
Appendix~\ref{app:robotics-results} gives the metric definitions and full breakdown.

\subsection{Implementation Details}
\label{sec:setup}

\paragraph{Driving model and optimization.}
All experiments are conducted on NVIDIA H100 GPUs. We use Cosmos~3-Nano (16B) \citep{agarwal2026cosmos3} with its language pathway frozen while training the video and action transformer. Images are stored at $832\times480$ resolution. We represent driving actions using controls rather than camera poses, as detailed in Appendix~\ref{app:data}. 

Driving ablations start from the same checkpoint and use a fresh optimizer for 10,000 training steps. The learning-rate schedule includes 1,000 warmup steps and cosine decay over the final 2,000 steps. For the ablations, we use two nodes with 16 GPUs, global batch size 16, and peak learning rate $10^{-5}$. For the full training, we use 8 nodes with a global batch size of 64. Evaluation uses nine history frames, the same text context, and 35 UniPC sampling steps \citep{zhao2023unipc}.

\begin{figure}[htbp]
\centering
\includegraphics[width=\linewidth]{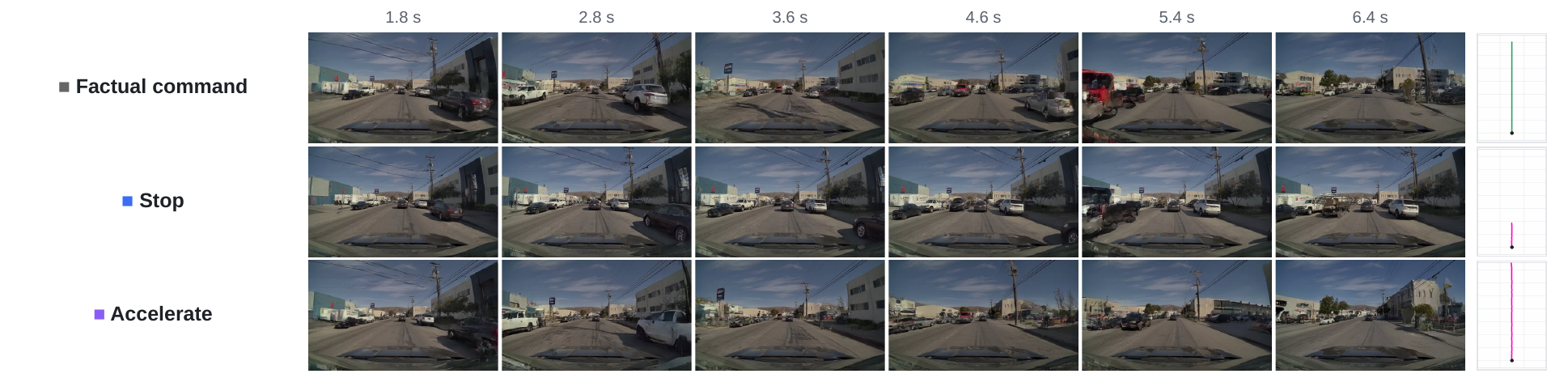}
\caption{Counterfactual ego control with \ourmethod. Rows show the generation under the factual ego, and stop, accelerate commands. Columns show corresponding times from 1.8 to 6.4\,s; the small plots at right show the ego trajectories associated with each row. 
}
\label{fig:counterfactual-ego}
\end{figure}

\textbf{Robotics model and generation.}
We use Cosmos~3-Nano \citep{agarwal2026cosmos3} as the backbone and fine-tune it with the bimanual action space from RoboTwin~\citep{chen2025robotwin}. To generate complete WorldArena episodes, the model produces video autoregressively in 32-frame chunks, using the final generated frame to condition the next chunk. Generation continues to the recorded episode length. Appendix~\ref{app:robotics-results} provides the action representations and training settings.

\begin{table}[tbp]
\caption{Driving evaluation under joint prediction and action-conditioned video generation. In Joint mode, Self ADE compares the IDM read-out of predicted video with the model's own action trajectory; Rollout ADE compares the generated action trajectory with the recording. In forward dynamics, Cmd ADE/FDE compare the IDM read-out with the supplied action command. \ourmethod-S and \ourmethod-M denote the ego-only and multi-agent model, respectively. }
\label{tab:main}
\centering
\small
\setlength{\tabcolsep}{3pt}
\begin{adjustbox}{max width=\linewidth}
\begin{tabular}{lcccccc}
\toprule
 & \multicolumn{3}{c}{Joint mode}& \multicolumn{3}{c}{Forward dynamics} \\
\cmidrule(lr){2-4}\cmidrule(lr){5-7}
Model & PSNR (dB) $\uparrow$ & Self ADE (m) $\downarrow$ & Rollout ADE (m) $\downarrow$ & PSNR (dB) $\uparrow$ & Cmd ADE (m) $\downarrow$ & Cmd FDE (m) $\downarrow$ \\
\midrule
Cosmos~3-Nano & 19.58 & 6.212 & 8.267 & 20.05 & 1.600 & 4.040 \\
\midrule
SFT Baseline & {20.38} & {0.538} & {2.920} & {21.23} & {1.023} & {2.803} \\
\ourmethod-S & {20.34} & {0.521} & {2.848} & {21.23} & {1.001} & {2.722} \\
\ourmethod-M & \textbf{20.42} & \textbf{0.513} & \textbf{2.804} & \textbf{21.28} & \textbf{0.981} & \textbf{2.671} \\
\bottomrule
\end{tabular}
\end{adjustbox}
\vspace{-0.5cm}
\end{table}

\subsection{Driving Results}
\label{sec:driving-results}

\subsubsection{Main driving comparison}
\label{sec:main}

\begin{wrapfigure}[18]{r}{0.42\textwidth}
\centering
\vspace{-0.5cm}
\includegraphics[width=\linewidth]{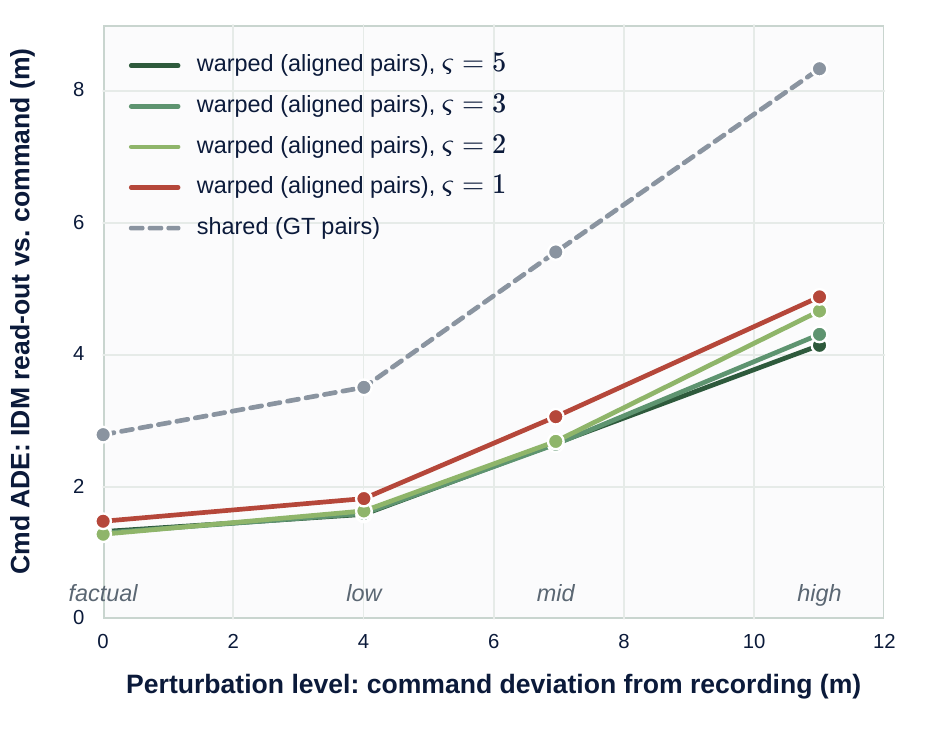}
\vspace{-0.5cm}
\caption{\small Ego-command counterfactuals on 100 clips. Each level is averaged over four commands (brake to stop, accelerate, nudge left, nudge right). Warped schedules ($\varsigma$ = 2, 3, 5) follow commands more closely than the shared schedule ($\varsigma$ = 1) at every level, with the gap widening as perturbations grow.}
\label{fig:counterfactual-ego-control}
\end{wrapfigure}

\textbf{Counterfactual ego control.}
Figure~\ref{fig:counterfactual-ego} shows responses to modified ego commands with the initial scene held fixed: stopping reduces forward viewpoint progression, acceleration advances its position relative to road boundaries and parked vehicles.
We report the quantitative results of different warp strength in Figure~\ref{fig:counterfactual-ego-control}. As the perturbation level increases, the Cmd ADE of both schedules increase. However, the warp schedule degrade less compared to the shared schedule with GT pairs, and consistently outperforming the one without aligned noising or warp schedule. 

\textbf{Command following and motion forecasting.} \ourmethod improves following of supplied ego commands relative to the SFT baseline (Table~\ref{tab:main}). In forward dynamics, both Cmd ADE and FDE decrease, indicating that video-inferred motion more closely matches the command. In Joint mode, Self ADE decreases from 0.538 to 0.513\,m, showing closer agreement between generated video and actions. Rollout ADE decreases from 2.920 to 2.804\,m, showing more accurate action forecasts against the recording. 

\subsubsection{Multi-agent controllability}
\label{sec:multiagent-results}

Our proposed multi-agent interface supports both predicting unspecified agent futures and conditioning video on supplied agent trajectories. To evaluate the controllability, we hold scene history, the ego command, and sampling noise fixed while changing a selected agent's future. In Table~\ref{tab:agent-controllability}, we input the generated videos and commanded trajectories to the Gemini 3.8 Flash model, asking it to rate the compliance on a scale of 0 to 1. This measures whether the intended agent follows the requested behavior in the generated video. Compared with the ego-only model, the multi-agent interface improves the mean compliance score from 10.07\% to 18.94\%, reporting the best performance over other methods,  because these models like Cosmos~3 Nano and the SFT baseline cannot receive control signal for other agents. 

\begin{table}[!htbp]
\begin{minipage}[t]{0.43\linewidth}
\caption{Counterfactual agent-command following. Scene history, the ego command, and sampling noise are held fixed while a selected agent’s future trajectory changes. Gemini 3.8 Flash evaluates agreement between the generated video and the supplied agent command; scores are reported on a percentage scale.}
\label{tab:agent-controllability}
\centering
\footnotesize
\setlength{\tabcolsep}{3pt}
\begin{adjustbox}{max width=\linewidth}
\begin{tabular}{@{}lc@{}}
\toprule
Model & VLM compliance (\%) $\uparrow$ \\
\midrule
Cosmos~3-Nano & 9.13 \\
SFT Baseline & 10.56 \\
\ourmethod-S & 10.07 \\
\ourmethod-M & 18.94 \\
\bottomrule
\end{tabular}
\end{adjustbox}
\end{minipage}\hfill
\begin{minipage}[t]{0.54\linewidth}
\vspace{0pt}
\centering
\includegraphics[width=\linewidth]{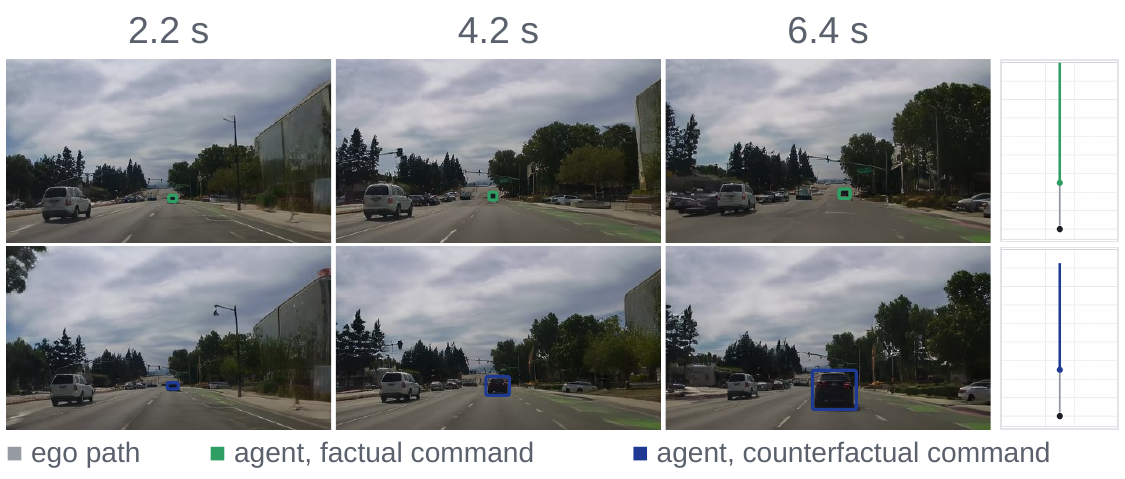}
\vspace{-16pt}
\makeatletter\def\@captype{figure}\makeatother
\caption{Counterfactual control of a surrounding vehicle. The top row uses the agent’s factual trajectory; the bottom row replaces it with a stop command. Columns show matching times, and the top-down view at right show the unchanged \textcolor[RGB]{150,155,162}{ego} path, \textcolor[RGB]{46,158,98}{factual} and \textcolor[RGB]{31,58,147}{modified} agent command. }
\label{fig:agent-control}
\end{minipage}
\end{table}

Figure~\ref{fig:agent-control} complements the scores with videos generated under matched factual and counterfactual commands. We stop the vehicle in front of the ego agent, and from the generated video, we can observe that the commanded agent clearly follows the input action. 
\begin{table}[thpb]
\caption{WorldArena Track-1 summary \citep{worldarena2025} on RoboTwin manipulation ($1{,}000$ episodes). Rows report six category averages and EWMScore, which is $100\times$ the mean of the 15 individual metrics.  Oracle evaluates ground-truth videos and is excluded from rankings. 
}
\label{tab:robo}
\centering
\small
\setlength{\tabcolsep}{5pt}
\begin{tabular}{lccccc}
\toprule
Category & {\textbf{Cosmos~3} \textbf{Zero-shot}} & {\textbf{Cosmos~3} \textbf{SFT}} & \textbf{GT-Src} & \textbf{CtrlWAM} & \textbf{Oracle} \\
\midrule
\textbf{Visual Quality} & 0.447 & 0.560 & \underline{0.563} & \textbf{0.576} & 0.598 \\
\textbf{Motion Quality} & 0.247 & \underline{0.316} & 0.306 & \textbf{0.327} & 0.348 \\
\textbf{Content Consistency} & 0.413 & \underline{0.650} & 0.620 & \textbf{0.665} & 0.662 \\
\textbf{Physics Adherence} & 0.258 & 0.456 & \underline{0.480} & \textbf{0.540} & 0.782 \\
\textbf{3D Accuracy} & 0.812 & 0.884 & \underline{0.899} & \textbf{0.921} & 0.956 \\
\textbf{Controllability} & 0.635 & 0.774 & \underline{0.788} & \textbf{0.819} & 0.869 \\
\midrule
\textbf{EWMScore} & 44.88 & \underline{58.70} & 58.65 & \textbf{61.75} & 66.92 \\
\bottomrule
\end{tabular}
\textranklegend
\end{table}

\subsection{Robotics Results}
\label{sec:robotics}

We evaluate whether aligned noising improves manipulation video generation under matched training budgets on the WorldArena benchmark~\citep{worldarena2025}.

\textbf{Training controls.}
The main comparison starts from a common checkpoint trained for 2,000 iterations. Each arm receives 2,000 additional iterations using aligned renders (\ourmethod), clean recorded frames in place of renders (GT-Src), or standard SFT of Cosmos~3-Nano.

\begin{figure}[!htbp]
\centering
\includegraphics[width=\linewidth]{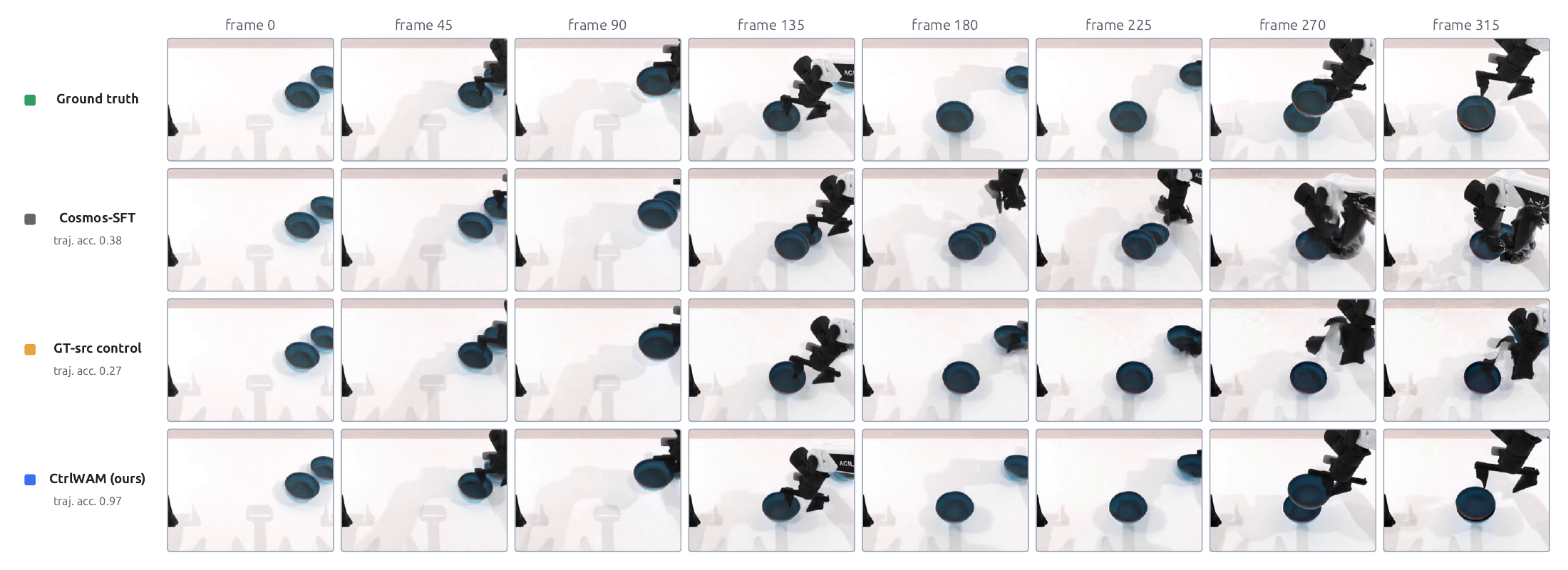}
\vspace{-0.5cm}
\caption{Qualitative RoboTwin manipulation comparison. Rows show the recorded video and generations from Cosmos-SFT, the GT-Src control, and CtrlWAM at matching frame indices. The trajectory-accuracy annotations apply to this episode, for which CtrlWAM more closely reproduces the recorded bowl motion and final arrangement. }
\label{fig:qualitative}
\end{figure}

Within the comparison in Tab.~\ref{tab:robo}, \ourmethod achieves the highest score in all six categories, with EWMScore 61.75 versus 58.70 for SFT and 58.65 for GT-Src.
Figure~\ref{fig:qualitative} compares a recorded bowl-manipulation sequence with generated videos at the same frame indices. In this example, \ourmethod more closely follows the recorded object motion and final configuration. 

\subsection{Ablations}
\label{sec:ablations}

\paragraph{Render source and video--action consistency}
\label{sec:render-source}

Table~\ref{tab:render} groups the eight-level Joint results by visual source, pairing shared and warped schedules within each group. GT-Src pairs perturbed actions with recorded future frames. Rendering uses simulator-generated frames directly, while Cleaned processes these frames with Difix3D+~\citep{difix3d2025} using the timestamp-paired recorded frames as a reference to remove the artifacts from the Gaussian renderings~\citep{alpasim_2025}.

\begin{table}[!htbp]
\begin{minipage}[t]{0.58\linewidth}
\caption{Joint-mode render-source comparison. Colors rank all six configurations within each metric.}
\label{tab:render}
\label{tab:abl-render8}
\end{minipage}\hfill
\begin{minipage}[t]{0.4\linewidth}
\caption{Joint-mode warp-strength sweep. When $\varsigma=1$, the model reduces to the shared schedule.}
\label{tab:abl-warp}
\end{minipage}
\par\smallskip
\begin{minipage}[t]{0.58\linewidth}
\centering
\footnotesize
\setlength{\tabcolsep}{2.5pt}
\begin{adjustbox}{max width=\linewidth}
\begin{tabular}{llccc}
\toprule
Source & Schedule & \shortstack{PSNR(dB) $\uparrow$}& \shortstack{Self ADE(m) $\downarrow$}& \shortstack{Rollout ADE(m) $\downarrow$}\\
\midrule
GT-Src & shared & 20.67 & \rankthird{0.845} & \ranksecond{2.968} \\
GT-Src & warped & 20.58 & 0.864 & 3.045 \\
\midrule
Rendering & shared & \ranksecond{21.21} & 0.916 & 3.153 \\
Rendering & warped & 20.87 & \ranksecond{0.798} & \rankthird{3.019} \\
\midrule
Cleaned & shared & \rankfirst{21.23} & 0.915 & 3.128 \\
Cleaned & warped & \rankthird{21.19} & \rankfirst{0.786} & \rankfirst{2.881} \\
\bottomrule
\end{tabular}
\end{adjustbox}
\resultranklegend
\end{minipage}\hfill
\begin{minipage}[t]{0.4\linewidth}
\centering
\footnotesize
\setlength{\tabcolsep}{2.5pt}
\begin{adjustbox}{max width=\linewidth}
\begin{tabular}{ccc}
\toprule
Warp $\varsigma$ & Self ADE (m) $\downarrow$ & Rollout ADE (m) $\downarrow$ \\
\midrule
1 & 0.916 & 3.153 \\
2 & 0.935 & {3.119} \\
3 & \underline{0.862} & \underline{3.081} \\
5 & \textbf{0.798} & \textbf{3.019} \\
\bottomrule
\end{tabular}
\end{adjustbox}
\textranklegend
\end{minipage}
\end{table}

The schedule comparison depends on the visual source. GT-Src performs better with the shared schedule, whereas Rendering and Cleaned favor the warped configuration for action errors. We believe that the warp mechanism delays the video denoising, not aligned with the original training schedule using GT pairs. For Rendering inputs, the warped configuration lowers these errors from 0.916/3.153 to 0.798/3.019\,m. Cleaned inputs show the largest reduction, from 0.915/3.128 to 0.786/2.881\,m, giving the lowest Self ADE and Rollout ADE among all six configurations.

Visual reconstruction instead favors shared scheduling for all three sources; the Cleaned configuration achieves the highest PSNR (21.23\,dB). This is the cost from the warped scheduling, because the videos are held on higher-level noises. However, the minor degradation in PSNR is almost negligible compared to the gains in Self ADE and Rollout ADE, as shown in Tab.~\ref{tab:render}.

\paragraph{Warped schedules improve video--action consistency}
\label{sec:schedule}

Table~\ref{tab:abl-warp} varies warp strength on Rendering inputs. Increasing $\varsigma$ from 1 to 5 lowers Self ADE from 0.916 to 0.798\,m, although the intermediate $\varsigma=2$ setting slightly worsens consistency. Warping therefore improves video--action consistency at selected strengths.

\section{Conclusion}
\label{sec:conclusion}

\ourmethod addresses intent--foresight misalignment by training on perturbed actions paired with their simulated visual consequences.
Warped video--action noise schedules aim to keep visual layout responsive while actions evolve, and multi-agent streams allow surrounding futures to be predicted or supplied as commands. Driving experiments show gains in command following, video--action consistency, and action forecasting; robotics controls support improved motion fidelity. Together, these findings support physically aligned denoising as a practical route toward world action models whose imagined outcomes better reflect their actions with improved controllability.

\textbf{Limitations.} The current model now inherits the bidirectional architecture of Cosmos~3, therefore, the generation latency is a bottleneck at this moment. We plan to distill the slow bidirectional model into a fast auto-regressive model and report the close-loop performance in future work.

\bibliography{iclr2027_conference}
\bibliographystyle{iclr2027_conference}

\input{sections/appendix}

\end{document}

%% file: math_commands.tex
\usepackage{amsmath,amsfonts,bm}

\def\eqref#1{equation~\ref{#1}}
\def\Eqref#1{Equation~\ref{#1}}

\def\1{\bm{1}}

\def\eps{{\epsilon}}

\def\vf{{\bm{f}}}

\DeclareMathAlphabet{\mathsfit}{\encodingdefault}{\sfdefault}{m}{sl}
\SetMathAlphabet{\mathsfit}{bold}{\encodingdefault}{\sfdefault}{bx}{n}



%% file: sections/appendix.tex
\DeclareRobustCommand{\appcheck}[2]{{\color[RGB]{170,65,0}\textsuperscript{\normalfont\scriptsize[#1]}#2}}
\clearpage
\appendix
\startcontents[appendix]
\section*{Appendix Contents}
\begingroup
\small
\printcontents[appendix]{}{1}{\setcounter{tocdepth}{2}}
\endgroup
\appendix

\section{Aligned denoising and optimization}
\label{app:aligned-noising}

\subsection{Paired states and corrective targets}
\label{app:standard-fm}
\label{sec:data}
\label{sec:objective}

With $\sigma=0$ denoting clean data, standard flow matching \citep{liu2023rectified,lipman2023flow} constructs $\x_\sigma=\sigma\eps+(1-\sigma)\x^{\mathrm{GT}}$ and regresses
\begin{equation}
\vf^{\mathrm{std}}=\eps-\x^{\mathrm{GT}}.
\label{eq:std-fm}
\end{equation}
Here $\x$ denotes either video or actions. The two modalities receive independent Gaussian draws; history tokens remain clean and are excluded from the loss.

For each recording, we independently sample Gaussian action noise and construct a perturbed command
\begin{equation}
\act_t^{\mathrm{cmd}}=t\act^\star+(1-t)\eps_a,
\qquad \eps_a\sim\mathcal{N}(0,I),\qquad \sigma_a=1-t.
\label{eq:cmd}
\end{equation}
The levels are predefined: $t\in\{0,0.25,0.5,0.75\}$ for the four-level grid, with $\{0.125,0.375,0.625,0.875\}$ added for eight levels. AlpaSim executes the noised commands in the reconstructed scene \citep{alpasim_2025}. We store these executed noised commands as $\act_t$, together with their rendered observations; the timestamp-paired recorded actions and video supply clean targets. Real history remains the conditioning prefix.

At training time, $\act_{\sigma_a}=\act_t$ is used without additional action noising. Given rendered video latents $\render$, fresh video noise produces $\vid_{\sigma_v}=\sigma_v\eps_v+(1-\sigma_v)\render$. Shared scheduling uses $\sigma_v=\sigma_a$; warped scheduling uses \Eqref{eq:ias-warp}. For either modality at its own $\sigma>0$, the target and clean-data estimate are
\begin{equation}
\vf^\ast=\frac{\x_\sigma-\x^{\mathrm{GT}}}{\sigma},
\qquad
\hat{\x}_0=\x_\sigma-\sigma\vf_\theta(\x_\sigma,\sigma,\mathrm{cond}).
\label{eq:target}
\end{equation}

\subsection{Loss weighting and recorded-data anchors}
\label{app:loss-weighting}
\label{sec:anchor}

Multiplying velocity MSE by the modality's squared noise level gives endpoint regression:
\begin{equation}
\mathcal{L}_{\x}=\sigma^2\|\vf_\theta-\vf^\ast\|^2
=\|\sigma\vf_\theta-(\x_\sigma-\x^{\mathrm{GT}})\|^2
=\|\hat{\x}_0-\x^{\mathrm{GT}}\|^2.
\label{eq:sigma-weight}
\end{equation}
Predictions and targets are scaled in fp32. This removes explicit low-noise amplification of endpoint error, without implying equal empirical gradients across levels. 

Each training window contributes one recorded video--action anchor trained with the original unweighted flow-matching objective and fresh Gaussian draws. Size-proportional sampling gives approximately 20\% anchors on four levels and 11\% on eight levels. Anchors retain exposure to real video and the continuous pretraining noise distribution. 

\subsection{Sampling}
\label{app:sampling}

Generation uses UniPC \citep{zhao2023unipc} in its data-prediction form. At each step the network velocity is converted to the clean estimate $\hat{\x}_0=\x_\sigma-\sigma\vf_\theta$ of \Eqref{eq:target}, and the multistep predictor--corrector update is formed from the history of these estimates.

\paragraph{Paired grids for warped checkpoints.}
The inference protocol that mirrors warped training runs the two modalities on distinct grids over the same raw times: the video block on the shift-$\varsigma$ grid, $\sigma_{v,k}=\mathrm{warp}(t_k;\varsigma)$, and the action block on the unshifted grid, $\sigma_{a,k}=t_k$, so that $\sigma_{v,k}=\mathrm{warp}(\sigma_{a,k};\varsigma)$ at every step, exactly as during training, and both reach $\sigma=0$ on the same step. Each step makes one joint network forward in which video tokens carry the embedding of
$\sigma_{v,k}$ and action tokens the embedding of $\sigma_{a,k}$. The predicted velocity is
split by block, and two independent UniPC solvers, each with its own multistep history,
update their block from their own clean estimate
$\hat{\x}_0=\x_{\sigma}-\sigma\vf_\theta$ at their own level.

\paragraph{Commands held fixed.}
In forward dynamics and agent control, the supplied future action tokens are marked by the
conditioning mask and enter the network at $\sigma=0$ with their clean values. They are
excluded from the noised state, their velocity is zero, and they are fixed points of the
solver, so they are never updated. Only the video tokens are integrated, conditioned on the
fixed command at every step. This is the mechanism that holds history tokens clean during
training, and it is how the counterfactual command experiments of Section~4.3 and
Appendix~\ref{app:ego-counterfactual-qualitative} are run.

\section{Driving data and experimental protocol}
\label{app:data}

\subsection{Data and aligned exports}
\label{app:driving-data}

PhysicalAI Autonomous Vehicles supplies the recordings \citep{nvidia2025physicalaiav}; its NuRec release supplies the reconstructed scenes \citep{nvidia2025physicalaiavnurec}. The export sweeps cover 8,924 scenes and 33,988 windows.  Validation and test exports contain clean rollouts and their paired recordings. Table~\ref{tab:data} gives the retained item counts; rejected renders mean that not every training window has every perturbation level.

\begin{table}[!htbp]
\caption{Training data inventory (training split unless noted).}
\label{tab:data}
\centering
\small
\begin{adjustbox}{max width=\linewidth}
\begin{tabular}{lrr}
\toprule
 & \textbf{Windows} & \textbf{Items} \\
\midrule
Aligned, 4-level grid ($\leq$4 levels/window) & 33{,}988 & 134{,}198 \\
Aligned, 8-level grid ($\leq$8 levels/window) & 33{,}988 & 269{,}151 \\
Ground-truth anchored        & 33{,}988 & 33{,}988 \\
Validation (real, pre-training set) & --- & 94{,}232 segments \\
Validation (simulation scenes, alternative) & 2{,}034 & 2{,}034 \\
Golden benchmark (real, standard driving evaluation) & 1{,}156 clips & 1{,}434 segments \\
\bottomrule
\end{tabular}
\end{adjustbox}
\end{table}

\paragraph{Export checks.}
Exports require complete camera streams, a uniform clock, 1.6\,s of real history, timestamp pairing, finite poses and actions, a $64\times100$\,ms action grid, and successful simulator decoding. Rollout-versus-dump disagreement must be below 5\,cm; action round-trip error must be below 0.5\,m and 0.05\,rad. 

Rendering uses simulator frames directly. Cleaned applies reference-conditioned Difix3D+ \citep{difix3d2025}.  Both variants share timestamps and the $832\times480$ resize operator. Validation uses real samples with the standard objective.

\subsection{Model inputs and initialization}
\label{app:driving-model}

\subsubsection{Packed inputs.}
The Cosmos~3-Nano backbone couples future video and action tokens through full attention while retaining causal attention on context tokens. Each input contains nine history and 32 future frames at 5\,Hz, with a 6.4\,s future horizon. The causal VAE compresses time by $4\times$ and space by $16\times$, giving 11 latent frames with 48 channels at $30\times52$ resolution; three latent frames encode history. The action stream contains 14 history rows derived from 16 recorded poses and 64 future rows at 0.1\,s intervals, padded to 64 channels. History is conditioned throughout; forward dynamics additionally conditions future actions.

\subsubsection{Unicycle actions.}
\label{sec:action_representation}
Each action row stores acceleration $a_t$ and curvature $\kappa_t$. Encoding fits smoothed heading to recorded yaw and speed to displacement projected along that heading, with initial speed estimated from history. Acceleration is the regularized finite difference of speed; curvature is $\Delta\theta_t/s_t$, where $s_t=v_t\Delta t+\tfrac12 a_t\Delta t^2$. Acceleration is standardized with mean $0.029$ and standard deviation $0.681$\,m/s$^2$; curvature uses mean $2.7\times10^{-4}$ and standard deviation $0.0261$\,m$^{-1}$. Feasibility bounds are $|a|\le9.8$\,m/s$^2$ and $|\kappa|\le0.33$\,m$^{-1}$. Only these two channels enter the noise, loss, and sampler; padding is masked. Decoding integrates from the recorded $t_0$ state. Counterfactual paths use the same encoding and decoding operators.

\subsubsection{Supervised initialization.}
\label{sec:pretraining}
The video and action experts are initialized from Cosmos~3-Nano~\citep{agarwal2026cosmos3} and trained, with the Wan~2.2~\citep{wan2025wan} VAE fixed. Four PhysicalAI-AV~\cite{nvidia2025physicalaiav} sources are sampled equally: chain-of-thought, chain-of-thought with meta-actions, 500k meta-action-only windows, and 500k trajectory-only windows. Joint, forward, and inverse tasks have probabilities 0.4/0.4/0.2. The pretraining uses 256 H100 GPUs, batch 256, and 20,000 steps; AdamW with $\beta=(0.9,0.999)$, $\epsilon=10^{-6}$, weight decay $10^{-5}$, gradient clipping 1.0, bf16 and fully sharded data parallelism; peak learning rate $1.5\times10^{-4}$ after 2,000 warmup steps, cosine decay to $10^{-7}$, and a $5\times$ action-input-projection multiplier. Noise levels are uniform and the objective is standard flow matching. The step-20,000 checkpoint initializes the driving post-training runs.

\subsection{Evaluation and main ablations}
\label{app:driving-protocol}
\label{app:results}

\subsubsection{Evaluation.}
The standard benchmark contains 1,156 clips and 1,434 segments with paired sampling noise. Self ADE compares video-inferred motion with the jointly predicted trajectory; Rollout ADE directly compares the predicted trajectory with the recorded GT trajectory. In forward dynamics, Cmd ADE/FDE compare the IDM read-out with the supplied command. IDM estimates the ego trajectory from the given video; Cmd FDE is computed from that same selected trajectory. The real-video IDM reference is 0.523/1.304\,m ADE/FDE. The 100-clip counterfactual evaluation in Appendix~\ref{app:ego-counterfactual-qualitative} is randomly selected from the held-out evaluation set.

\subsubsection{Optimization.}
The main driving ablations start from the same pretrained checkpoint and use a fresh optimizer for 10,000 steps, 16 H100 GPUs, global batch 16, and peak learning rate $10^{-5}$. The schedule uses 1,000 warmup steps and cosine decay over the final 2,000 steps. Full ego-only and multi-agent training uses the matched eight-node, batch-64 recipe in Appendix~\ref{app:agents}.

\subsubsection{Schedule-comparison settings.}
\label{app:warp-sweep-protocol}
Table~\ref{tab:render} compares render sources under different schedules. For warped rows we use $\varsigma=5$.  Table~\ref{tab:abl-warp} is conducted on the Rendering sources (not cleaned by Difix3D~\citep{difix3d2025} model) and varies $\varsigma\in\{1,2,3,5\}$, with the same initialization, step count, batch size, and learning rate.

\section{Counterfactual ego-control protocol and examples}
\label{app:ego-counterfactual-qualitative}

Figure~\ref{fig:counterfactual-ego-control} evaluates 100 randomly sampled held-out clips. History frames, text, and sampling noise are held fixed across command variants. Commands use the training-time unicycle encoding, and Cmd ADE compares the IDM read-out with the supplied command.

\begin{figure}[!htbp]
\centering
\includegraphics[width=\linewidth]{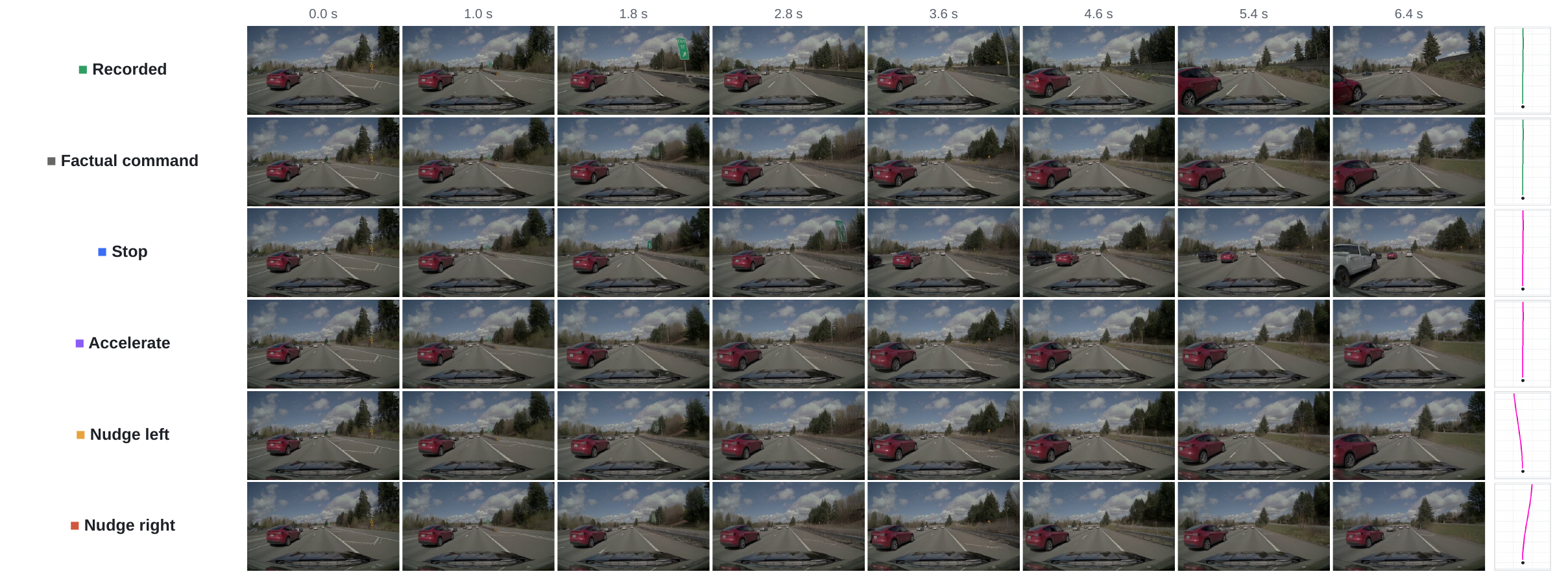}
\par\smallskip
\includegraphics[width=\linewidth]{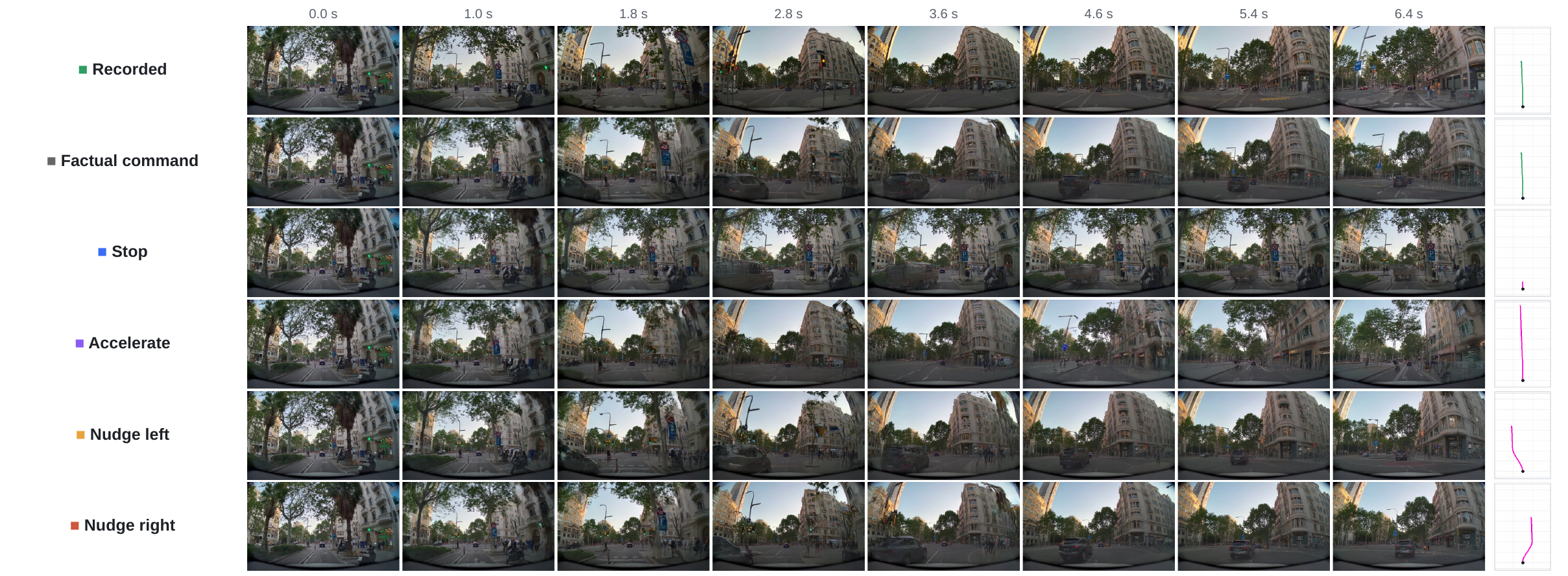}
\caption{Additional counterfactual ego control on a highway (top) and at an urban intersection (bottom). Each panel compares the recording with generations under factual, stop, accelerate, nudge-left, and nudge-right commands at matching times over 0.0--6.4\,s. Plots at right show the corresponding ego trajectories.}
\label{fig:counterfactual-extra}
\end{figure}

Figure~\ref{fig:counterfactual-extra} adds highway and intersection examples to Figure~\ref{fig:counterfactual-ego}; these illustrate command-dependent changes rather than an additional aggregate result.

\paragraph{Perturbation protocol.}
Each counterfactual replaces the recorded ego trajectory with one of four edited commands: \emph{brake to stop} decelerates at a constant rate until the ego halts; \emph{accelerate} increases speed at a constant rate; and \emph{nudge left/right} offsets the recorded path laterally by a distance. Longitudinal strengths are in m/s$^2$ and lateral strengths in metres, so severity is defined by pairing one strength per type. The low/mid/high settings use braking decelerations of $1/2/3$\,m/s$^2$, accelerations of $0.75/1.5/3$\,m/s$^2$, and left/right offsets of $2/3.5/5$\,m, respectively.

At each severity level, we average Cmd ADE over the four command types. The common horizontal axis is the mean ADE between the edited command and the recording, identical for every checkpoint: 4.0\,m (low), 7.0\,m (mid), and 11.0\,m (high). The factual recorded command is the leftmost point at 0\,m deviation.

\section{Robotics protocol and metric breakdown}
\label{app:robotics-results}

\subsection{Training and generation}

The main comparison uses 14-dimensional absolute joint positions (12 arm joints and two grippers) in 32-step chunks. A common joint-space SFT checkpoint trained for 2,000 iterations initializes three further 2,000-iteration stages: continued SFT, GT-Src, and aligned post-training. Each stage restarts the optimizer and schedule. GT-Src replaces aligned rendered frames with real video while retaining the perturbed actions and aligned recipe. Each aligned stage uses 59,696 simulator-derived samples and a $6{:}2$ aligned-to-anchor mixture.

An aligned sample pairs its conditioning frame and 32 simulator frames with the executed noised command $\x_\sigma$, at $\sigma=1-t_{\mathrm{clean}}$ and $t_{\mathrm{clean}}\in\{0,0.25,0.5,0.75\}$. A Gaussian draw sampled independently per window ; the clean rollout is rendered as a validity gate. Action states enter verbatim, while video latents receive fresh Gaussian noise. Both modalities use fp32 endpoint supervision with conditioning rows masked; anchors use the standard objective. The SFT loss scale and action-loss weight remain 10.

Training uses eight H100 GPUs, FusedAdam, nominal learning rate $10^{-4}$, weight decay 0.05, and a $5\times$ action-projection multiplier. A linear multiplier decays from 0.4 to zero over each 2,000-iteration stage without warmup. Token packing uses 45,056 tokens per rank, approximately 11 samples per rank and global batch 88. Generation uses forward dynamics with 30 UniPC steps, shift 10, guidance 1, and seed 0. The last generated frame conditions the next chunk until the recorded episode length is reached. Robotics output is $736\times544$ at 30\,Hz; the fixed 6.4\,s horizon and $832\times480$ resolution apply to driving.

\subsection{WorldArena evaluation}

Evaluation uses 1,000 held-out RoboTwin episodes and the WorldArena harness \citep{worldarena2025}. Table~\ref{tab:robo-post} consolidates the individual metrics for the models in Table~\ref{tab:robo}. Visual and consistency metrics build on VBench \citep{huang2024vbench}; Flow Score follows EvalCrafter \citep{liu2024evalcrafter}; motion smoothness and photometric consistency follow WorldScore \citep{duan2025worldscore}. JEPA similarity follows JEDi \citep{luo2024jedi}, and trajectory accuracy uses the normalized dynamic-time-warping comparison adopted by EWMBench \citep{yue2025ewmbench}. Other metric definitions and normalization follow WorldArena. Flow, trajectory, photometric, and motion scores use the harness's published percentile-based min--max bounds. EWMScore is 100 times the mean of the 15 normalized metrics. 

\begin{table}[!htbp]
\caption{Individual WorldArena metrics for Table~\ref{tab:robo}. SFT, GT-Src, and \ourmethod share a joint-space SFT 2k initialization and receive 2k additional iterations. Zero-shot uses text conditioning; Oracle scores recorded videos. EWMScore is $100$ times the mean of 15 normalized metrics. Oracle JEPA and depth ($^*$) are identity-comparison checks; photometric consistency ($^\dagger$) can favor smoother-than-real motion.}
\label{tab:robo-post}
\centering
\small
\setlength{\tabcolsep}{4pt}
\begin{adjustbox}{max width=\linewidth}
\begin{tabular}{lccccc}
\toprule
Metric & Zero-shot & SFT & GT-Src & \ourmethod & Oracle \\
\midrule
Image Quality & 0.398 & 0.396 & 0.394 & 0.394 & 0.401 \\
Aesthetic Quality & 0.387 & 0.377 & 0.375 & 0.371 & 0.396 \\
JEPA Similarity & 0.555 & 0.905 & 0.921 & 0.963 & 0.998$^{*}$ \\
Dynamic Degree & 0.142 & 0.199 & 0.190 & 0.212 & 0.229 \\
Flow Score & 0.073 & 0.099 & 0.095 & 0.109 & 0.125 \\
Motion Smoothness & 0.525 & 0.651 & 0.634 & 0.660 & 0.690 \\
Subject Consistency & 0.440 & 0.711 & 0.667 & 0.737 & 0.753 \\
Background Consistency & 0.475 & 0.762 & 0.714 & 0.785 & 0.811 \\
Photometric Consistency$^{\dagger}$ & 0.325 & 0.476 & 0.478 & 0.474 & 0.423 \\
Interaction Quality (VLM) & 0.417 & 0.611 & 0.633 & 0.668 & 0.719 \\
Trajectory Accuracy & 0.099 & 0.302 & 0.327 & 0.412 & 0.845 \\
Depth Accuracy & 0.772 & 0.918 & 0.930 & 0.955 & 1.000$^{*}$ \\
Perspectivity (VLM) & 0.853 & 0.850 & 0.867 & 0.887 & 0.911 \\
Instruction Following (VLM) & 0.405 & 0.651 & 0.678 & 0.738 & 0.830 \\
Semantic Alignment & 0.865 & 0.897 & 0.897 & 0.899 & 0.907 \\
\textbf{EWMScore} (mean of 15 $\times$ 100) & 44.88 & 58.70 & 58.65 & 61.75 & 66.92 \\
\bottomrule
\end{tabular}
\end{adjustbox}
\end{table}

Individual metrics do not uniformly favor aligned training: continued SFT has higher image and aesthetic quality. Aggregate improvements therefore reflect gains across several measures rather than dominance on every metric. Ground-truth oracle scores are calibration references rather than strict upper bounds, because the GT videos provided by WorldArena~\citep{worldarena2025} is of lower resolution than the one used in our experiments. 

\section{Selected-agent interface and evaluation}
\label{app:agents}

\subsection{Representation and selection}

For each scene, the interface contains an ego block and $K$ selected surrounding-agent blocks, with $0\le K\le8$ varying across scenes:
\begin{equation}
\act=[\act^{(0)}\Vert\act^{(1)}\Vert\cdots\Vert\act^{(K)}],
\qquad
\act^{(k)}=[\act^{(k)}_{\mathrm{hist}}\Vert\act^{(k)}_{\mathrm{fut}}].
\label{eq:agent-rows}
\end{equation}
Each block contains 14 history and 64 future rows at 0.1\,s intervals. Only selected agents receive explicit action blocks and trajectory supervision. Histories are clean context; all selected futures are prediction targets in Joint mode, or clean commands when supplied for controlled generation. With $K=0$, the input reduces to the ego-only layout.

Agent histories contain ego-frame $(x,y)$ positions at $t_0$, scaled by 20\,m, while future unicycle controls use each actor's local $t_0$ frame. Predicted trajectories are transformed to the recorded trajectories' frame for evaluation. Tracks require finite states throughout the horizon, feasible controls, and an encode--decode error below 0.5\,m. Missing leading history is back-filled from the earliest finite state.

\paragraph{Agent selection and encoding .}
Static actors with net displacement below 2\,m are excluded before packing. Remaining actors must project into the front image at positive depth, lie within 60\,m, span at least 12\,px in height at $832\times480$, and pass the projected-occlusion test (a nearer retained box contains the candidate centre with a 1\,m depth margin). The nearest eligible actors up to $K_{\max}=8$ are retained. Actor tokens use a separate embodiment projection initialized from the ego projection, and spatial rotary coordinates given by the actor's projected $t_0$ position on the $15\times26$ latent grid. Temporal rotary positions match the ego block. For $K>0$, ego future rows are weighted by $1+K$ and each selected-agent block by $(1+K)/K$; for $K=0$, no agent loss is computed.

\subsection{Matched training and evaluation}
We first pretrain the model as described in Sec.~\ref{sec:pretraining}. 
\ourmethod-S and \ourmethod-M use the same initialization, training data, aligned-noising recipe, training duration, optimizer, learning-rate schedule, batch size, and evaluation protocol. They differ only in the additional agent interface. Both start from the pretrained checkpoint and use eight nodes, batch 64, learning rate $3\times10^{-5}$ for 20k steps.  For a fair comparison, we continue to finetune the pretrained checkpoints using the standard noising with the same setting, reporting the SFT baseline results. The Agent ADE/FDE of 3.215/7.224\,m compare generated trajectories directly with recordings: average within each evaluation batch over selected actors, then average batches with batch-size weights. Under selection, 4,084 actors in 1,197 of the 1,434 benchmark segments enter these metrics. Agent prediction errors are not evaluated in forward dynamics, where future actions are supplied, because we do not have an inverse dynamic models (IDM) to estimate the agents' trajectories from the given video. 

\paragraph{Counterfactual actor control.}
\label{app:agent-control-evaluation}
For Table~\ref{tab:agent-controllability}, scene history, the ego command, and sampling noise are held fixed while the selected actor's future changes. \ourmethod-M receives the edited agent command through its agent interface. Cosmos~3 zero-shot, the SFT baseline, and \ourmethod-S receive no non-ego agent command; their scores measure agreement with the requested agent behavior without conditioning on that command.

Gemini 3.8 Flash receives the generated video and commanded trajectory and assigns a compliance score $s_i\in[0,1]$. The reported percentage is $100\,N^{-1}\sum_{i=1}^{N}s_i$, where $N$ is the number of scored examples. We use 100 clips for evaluation, each with four type of counterfactual commands. This is a mean judge score, with no thresholding into success or failure. 

\section{Discussion and limitations}
\label{app:discussion}

The experiments assess generated futures and offline open-loop metrics. Self ADE measures video--action consistency, Rollout ADE measures direct forecasting accuracy, and Cmd ADE/FDE measure agreement with supplied commands through an IDM. Most robotics metrics also rely on external models. Both the evaluation is heavily dependent on the used models, leading to a risk of the model bias.

\textbf{Limitations.}
The aligned noising pipeline requires a simulator to execute perturbed actions and render their consequences. The multi-agent extension is limited to driving because we lack suitable dynamic robotics datasets with actor-level action--observation pairs. We expect future real--sim--real workflows to enable reconstruction of multi-agent interactions, generation of controlled variations, and transfer of learned behavior back to real environments, making multi-agent prediction and control in robotics feasible.